\documentclass[runningheads]{llncs}
\usepackage[T1]{fontenc}
\usepackage{amsmath}
\usepackage{graphicx}
\usepackage{tabularx}

\begin{document}
\title{Resource-Efficient Glioma Segmentation on Sub-Saharan MRI}
%
%

\author{Freedmore Sidume\inst{1*}\orcidID{0000-0002-4470-4814} \and
Oumayma Soula\inst{2,3*}\orcidID{0009-0008-9576-7285} \and
Joseph Muthui Wacira\inst{4*}\orcidID{0000-0002-9544-2324} \and
YunFei Zhu\inst{5}\orcidID{0009-0004-0590-3150} \and
Abbas Rabiu Muhammad\inst{6,7}\orcidID{0000--0002-7139-9341} \and
Abderrazek Zeraii\inst{2}\orcidID{000-0002-0697-0863}\and 
Oluwaseun Kalejaye\inst{8}\orcidID{0009-0007-9934-109X}\and
Hajer Ibrahim \inst{2,9}\orcidID{0009-0002-8192-8777} \and
Olfa Gaddour \inst{10}\orcidID{0000-0002-1062-5528} \and 
Brain Halubanza \inst{11}\orcidID{0000--0002-7756-7609} \and
Dong Zhang\inst{12,13}\orcidID{0000-0002-2948-1384} \
Udunna C Anazodo\inst{5,12,13,14,15,16,17}\orcidID{0000-0001-8864-035X}
Confidence Raymond \inst{5,12,14}\orcidID{0000-0003-3927-9697}}

\authorrunning{F. Sidume et al.}
%
\institute{BAC School of Computing and Information Systems, Botswana \and
Faculty of Medicine of Sfax, University of Sfax, Sfax, Tunisia \and
Laboratory of Biophysics and Medical Technologies, 
Higher Institute of Medical Technologies of Tunis, University ElManar, Tunis, Tunisia \and 
London School of Hygiene and Tropical Medicine, London, UK \and
Lawson Health Research Institute, London, Ontario, Canada \and
Department of Radiology, Bayero University, Kano-Nigeria \and
Department of Radiology, Aminu Kano Teaching Hospital, Kano-Nigeria \and
Department of Anatomy,
Federal University Lokoja college of health sciences, Kogi State, Nigeria\and
Department of Computer Science and Information, College of Science, Majmaah University, AL-Majmaah, 11952, Saudi Arabia\and
CES Laboratory, National School of Engineers of Sfax, Sfax, Tunisia \and
Department of Computer Science and IT, Mulungushi University, Zambia \and
Multimodal Imaging of Neurodegenerative Diseases (MiND) Lab, Department of Neurology and Neurosurgery, McGill University, Montreal, QC, Canada\and
Department of Electrical and Computer Engineering, University of British Columbia, Vancouver, Canada\and
Medical Artificial Intelligence Laboratory (MAI Lab), Lagos, Nigeria  \and
Montreal Neurological Institute, McGill University, Montréal, Canada \and
Department of Medicine, University of Cape Town, South Africa \and
Department of Clinical \& Radiation Oncology, University of Cape Town, South Africa
}

\renewcommand{\thefootnote}{\fnsymbol{footnote}}
\footnotetext[1]{These authors contributed equally to this work.}
\renewcommand{\thefootnote}{\arabic{footnote}}
\maketitle              
\newpage
\begin{abstract} 
Gliomas are the most prevalent type of primary brain tumors, and their accurate
segmentation from MRI is critical for diagnosis, treatment planning, and longitudinal monitoring.
However, the scarcity of high-quality annotated imaging data in Sub-Saharan Africa (SSA) poses
a significant challenge for deploying advanced segmentation models in clinical workflows. This
study introduces a robust and computationally efficient deep learning framework tailored for
resource-constrained settings. We leveraged a 3D Attention U-Net architecture augmented with
residual blocks and enhanced through transfer learning from pre-trained weights on the BraTS 2021
dataset. Our model was evaluated on 95 MRI cases from the BraTS-Africa dataset, a benchmark for
glioma segmentation in SSA MRI data. Despite the limited data quality and quantity, our approach
achieved Dice scores of 0.76 (Enhancing Tumor - ET), 0.80 (Necrotic and Non-Enhancing Tumor
Core - NETC), and 0.85 (Surrounding Non-Functional Hemisphere - SNFH). These results
demonstrate the generalizability of the proposed model and its potential to support clinical decision
making in low-resource settings. The compact architecture ($\sim 90$ MB) and sub-minute per-volume
inference time on consumer- grade hardware, further underscores its practicality for deployment in
SSA health systems. This work contributes toward closing the gap in equitable AI for global health
by empowering underserved regions with high performing and accessible medical imaging solutions.

\end{abstract}

\section{Introduction}

Brain tumors are a heterogeneous group of neoplasms that include gliomas, meningiomas, and pituitary adenomas, and can severely affect cognition, motor function, and quality of life. Gliomas are the most aggressive and account for the majority of malignant brain tumors in the central nervous system, with nearly 80\% of patients dying within two years of diagnosis \cite{baid_rsna-asnr-miccai_2021}. Magnetic resonance imaging (MRI) is central to glioma diagnosis and treatment planning, providing high-resolution, multi-parametric images across modalities such as T1-weighted (T1w), contrast-enhanced T1w (T1CE), T2-weighted (T2w), and FLAIR, which together capture tumor location, infiltration, and morphology \cite{sahayam_integrating_2024}. However, accurate delineation still relies on expert radiologists, who are scarce in resource-limited settings such as sub-Saharan Africa (SSA) \cite{nigatu_medical_2023,murali_bringing_2024}. Manual interpretation is further complicated by the large data volume, multiple modalities, and frequent presence of low-quality images.

Deep learning (DL) methods, particularly convolutional neural networks \\(CNNs), have advanced automated brain tumor segmentation by capturing complex spatial patterns and providing accurate delineation of tumor subregions. Architectures such as DeepMedic \cite{kamnitsas_efficient_2017} and 3D U-Net have achieved state-of-the-art performance, while transfer learning has improved generalization on smaller or domain-specific datasets \cite{tajbakhsh_convolutional_2016}. Yet, most approaches require substantial computational resources, limiting deployment in low- and middle-income countries (LMICs).

The Brain Tumor Segmentation (BraTS) Challenge, established in 2012, has driven progress by providing annotated datasets and standardized benchmarks. However, models trained on BraTS data often underperform in SSA due to differences in imaging protocols, resolution, and annotation scarcity \cite{zhang2022stroke}. Recent initiatives, such as BraTS-Africa \cite{adewole_brain_2023}, address this gap by curating SSA-specific datasets. Studies have shown that fine-tuning pre-trained models on local data improves segmentation, underscoring the need for domain adaptation \cite{barakat_towards_2023,amod_bridging_2023}.

Building on these efforts, we propose \textbf{BrainUNet}, a lightweight 3D U-Net model with residual connections and attention mechanisms. Pre-trained on the BraTS 2021 dataset \cite{baid_rsna-asnr-miccai_2021} and fine-tuned on BraTS-Africa \cite{adewole_brain_2023}, BrainUNet is optimized for the challenges of SSA imaging, including low resolution and modality variation. Its efficient design balances accuracy and computational cost, making it suitable for deployment in resource-constrained clinical environments.

\section{METHODS}
\subsection{The Dataset}

The BraTS-Africa dataset comprises multi-parametric MRI (mpMRI) scans from 95 glioma patients (60 training, 35 validation) \cite{adewole_brain_2023}. Each case includes four structural modalities: native T1-weighted (T1w), post-contrast T1-weighted (T1CE), T2-weighted (T2w), and T2 Fluid Attenuated Inversion Recovery (FLAIR). In this study, the native T1w modality was excluded. All scans were reviewed by board-certified neuroradiologists, pre-processed, and manually annotated using standardized BraTS protocols. Ground-truth tumor subregions—Enhancing Tumor (ET), Non-enhancing Tumor Core (NETC), and Surrounding Non-enhancing FLAIR Hyperintensity (SNFH)—were delineated and verified by expert neuroradiologists in accordance with BraTS 2023 challenge criteria \cite{baid_rsna-asnr-miccai_2021,adewole_brain_2023}.

\subsection{Proposed Approach}

\subsection{Proposed BrainUNet Framework}
\begin{figure}[h!]
    \centering
    \includegraphics[width=0.7\textwidth]{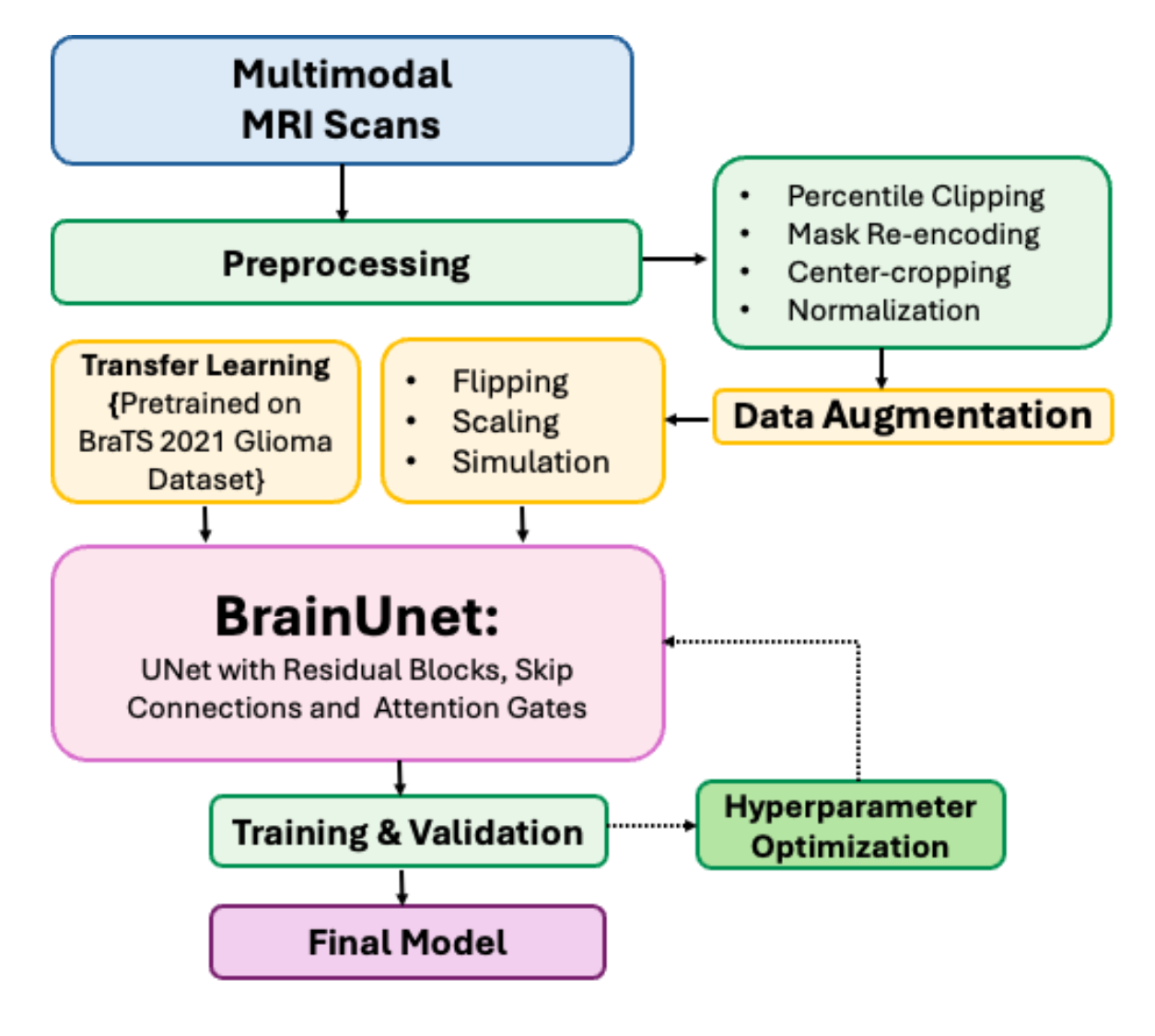}
    \caption{Overview of the proposed BrainUNet framework.}
    \label{fig:approach}
\end{figure}

\begin{figure}[h!]
    \centering
    \includegraphics[width=0.9\textwidth]{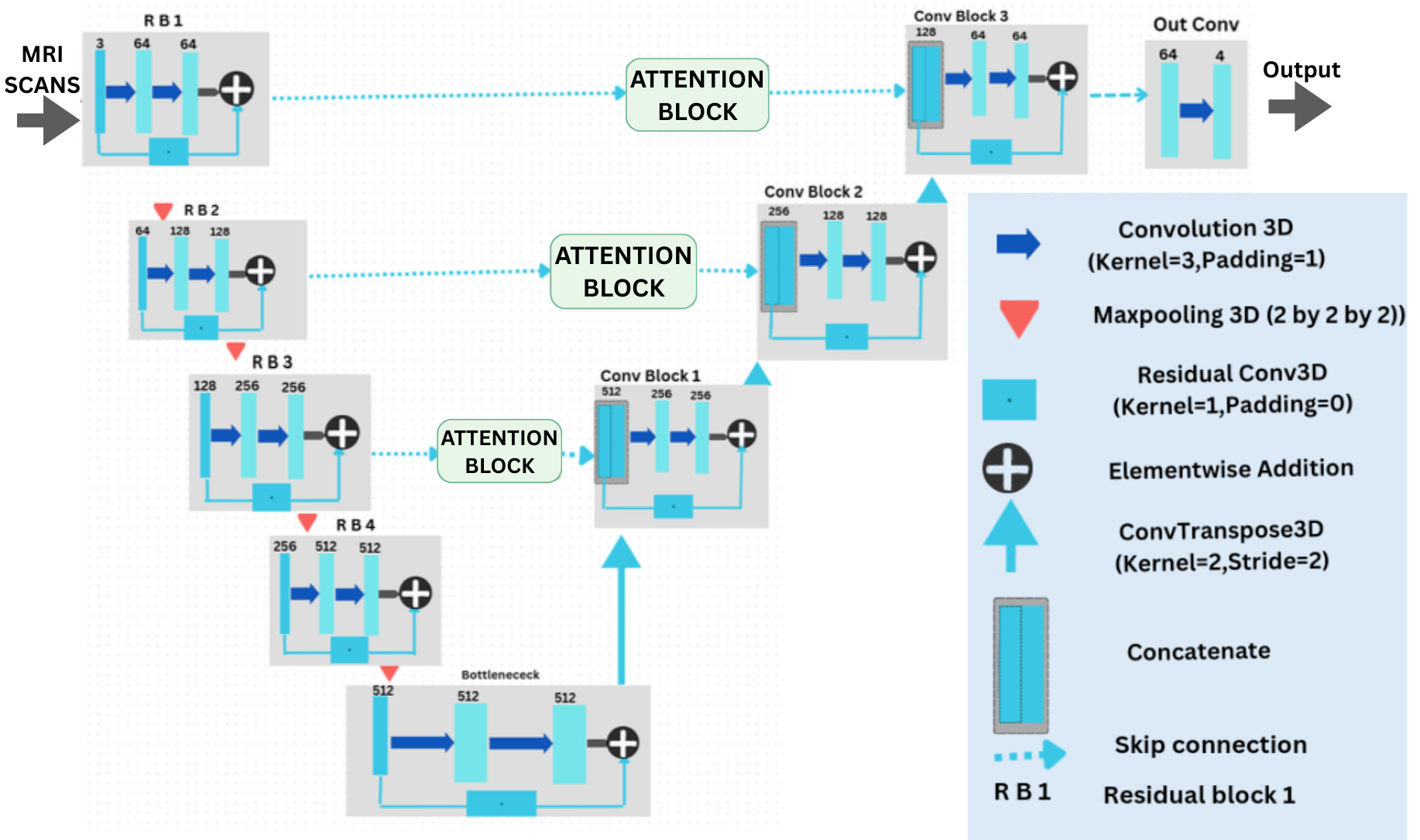}
    \caption{3D Attention U-Net with residual connections.}
    \label{fig:model}
\end{figure}

We propose \textbf{BrainUNet}, a 3D Attention U-Net with residual blocks as its backbone (Figures~\ref{fig:approach}, \ref{fig:model}). The framework integrates transfer learning with tailored preprocessing to enhance MRI quality and minimize information loss. The U-shaped encoder–decoder structure extracts multi-scale features while skip connections preserve spatial detail. Attention gates at skip connections focus the network on clinically relevant regions, improving tissue discrimination critical for diagnosis \cite{hassanin_visual_2024}. Residual connections in both encoder and decoder blocks stabilize training, reduce computational complexity, and enhance feature representation.  


\subsubsection{Data Preprocessing}
MRI scans were cropped to $128 \times 128 \times 128$ to reduce computational demand. Three modalities (FLAIR, T1CE, T2W) were stacked, providing optimal tumor contrast \cite{abidin_recent_2024}. Percentile clipping mitigated outliers, while intensity normalization harmonized modalities and masks \cite{ranjbarzadeh_brain_2021}. These steps improved efficiency and preserved critical information for robust segmentation.  

\subsubsection{Data Augmentation}
To enhance generalization on the BraTS-Africa dataset, we applied augmentations including flipping, scaling, and gamma adjustment to account for variability in orientation and brightness. Motion and ghosting artifacts were introduced to simulate common MRI imperfections, improving robustness to real-world clinical imaging noise \cite{sajjad_multi-grade_2019}.  

\subsubsection{The model design:}
\subsection{Model Architecture}
\begin{figure}[!h]
    \centering
    \includegraphics[width=4.5cm]{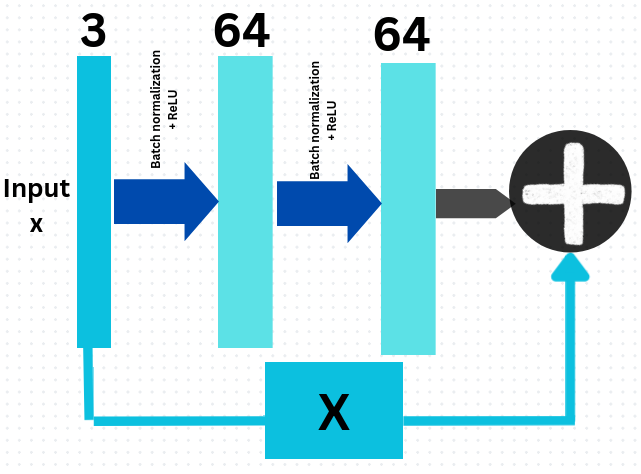} 
    \quad
    \includegraphics[width=6.5cm]{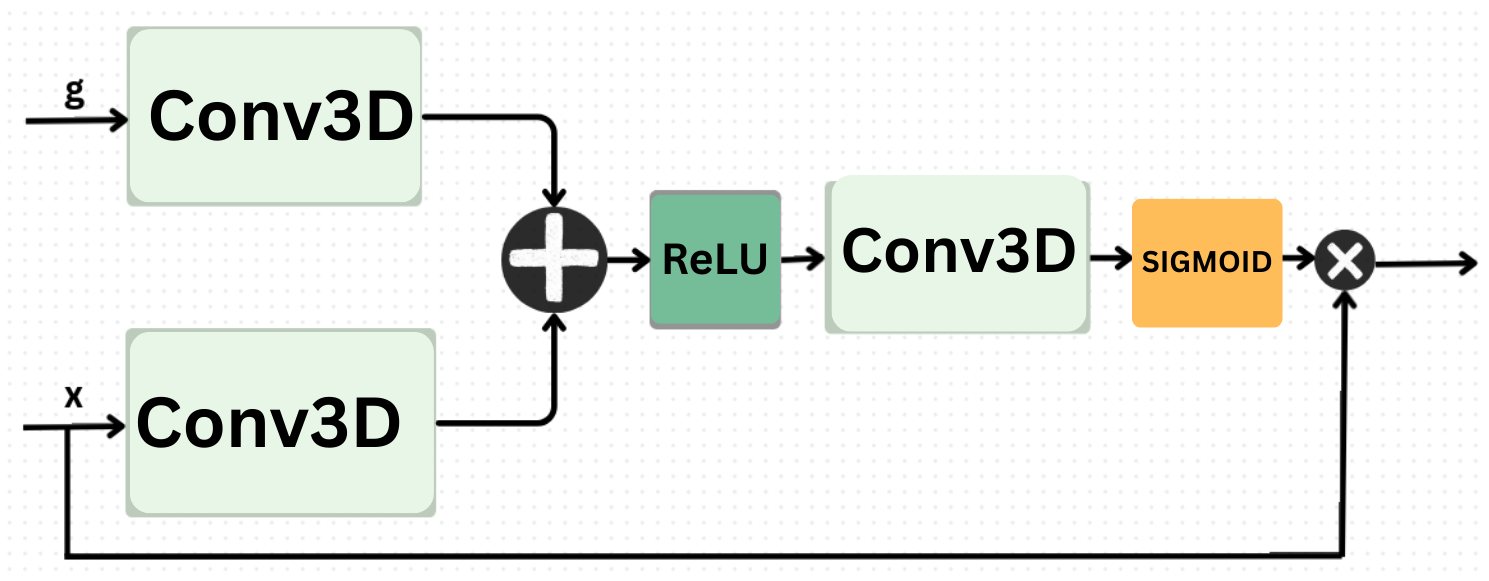}
    \caption{Core components of BrainUNet: (a) residual block with skip connections, and (b) attention block with gating for region-specific focus.}
    \label{fig:blocks}
\end{figure}

The proposed BrainUNet architecture (Figure~\ref{fig:model}) builds on the U-Net framework to improve 3D medical image segmentation. It integrates two key components designed to enhance learning and focus within volumetric data:  

\begin{enumerate}
    \item \textbf{Residual Blocks}:  
    Instead of standard convolutional blocks, the model employs residual blocks (Figure \ref{fig:blocks}a), each consisting of two 3D convolutional layers, ReLU activation, batch normalization, and a skip connection. Residual connections mitigate the vanishing gradient problem, preserve identity information, and facilitate deeper network training. This improves convergence speed, feature learning, and the ability to capture complex patterns.  

    \item \textbf{Attention Mechanisms}:  
    Attention blocks (Figure \ref{fig:blocks}b) enable the network to focus on salient regions by adaptively re-weighting feature maps. Each block receives $x$ (encoder features via skip connections) and $g$ (decoder features from deeper layers), which are processed through 3D convolutions, batch normalization, and activations. The gating mechanism emphasizes informative regions, improving discrimination of subtle tissue differences critical for clinical decision-making.  
\end{enumerate}

\vspace{-25pt}

\subsubsection{Transfer Learning:}
Before fine-tuning on BraTS-Africa, the model was pre-trained on the larger BraTS-GLI 2021 dataset ($n=1251$ 3D MRI scans). Training used the parameters in Table \ref{tab:parameters}, with epochs and learning rate set to 30 and $1 \times 10^{-3}$, respectively. This pre-training phase exposed the model to diverse tumor types, shapes, and stages, enabling it to learn robust, generalized features while mitigating overfitting. Such a foundation improved transferability and provided a strong basis for fine-tuning on BraTS-Africa, enhancing clinical applicability.

\subsubsection{Training and validation:}
The pre-trained model was fine-tuned using 50 training and 10 validation samples. The hyperparameters are summarized in Table \ref{tab:parameters}.  

\begin{table}[!h]
\centering
\begin{tabularx}{0.7 \textwidth} { 
  | >{\raggedright\arraybackslash}X 
  | >{\centering\arraybackslash}X | }
  \hline
  \textbf{Parameter} & \textbf{Choice}\\
  \hline
  Loss & Tversky loss \cite{salehi_tversky_2017}\\
 \hline
 Optimizer & Adam \cite{kurbiel_training_2017} \\
 \hline
 Learning rate  & $1 \times 10^{-4}$ \\
\hline
 Epochs & 50\\
\hline
\end{tabularx}
\vspace{5pt}
\caption{Fine-tuning hyperparameters.}
\label{tab:parameters}
\end{table}

\vspace{-25pt}

\footnotetext{\textit{https://github.com/CAMERA-MRI/SPARK2024/tree/main/BrainUNet}}
\paragraph{Rationale.}
Adam optimizer combines the advantages of AdaGrad and RMSProp, enabling efficient training with adaptive learning rates \cite{kurbiel_training_2017,mzoughi_deep_2020}. Tversky loss addresses class imbalance by weighting false positives and false negatives through $\alpha$ and $\beta$. It is defined as:  
\vspace{-5pt}
\begin{equation}
\text{Tversky Loss} = 1 - \text{TI}, \quad 
\text{TI} = \frac{\sum_{i} y_i \hat{y}_i}{\sum_{i} y_i \hat{y}_i + \alpha \sum_{i} \hat{y}_i (1-y_i) + \beta \sum_{i} y_i (1-\hat{y}_i)}
\end{equation}

where $y_i$ and $\hat{y}_i$ denote ground-truth and predicted labels. This formulation improves robustness by emphasizing minority tumor classes.

\subsubsection{Evaluation Metrics:} 

The BrainUNet model was evaluated on the BraTS-Africa dataset ($n=95$) using 5-fold cross-validation to ensure robustness and mitigate bias from limited data. For each fold, one partition served as validation ($n=35$) while the remainder was used for training, and results were averaged across folds. Performance was assessed using Dice score, Intersection over Union (IoU), and Hausdorff95 distance, tracked over 100 epochs. Predicted tumor sub-regions from the validation set were further benchmarked on the BraTS Challenge Synapse platform \url{www.synapse.org/brats}, with and without fine-tuning, to obtain standardized metrics \cite{baid_rsna-asnr-miccai_2021}. Finally, runtime efficiency was measured by model size (91.4 MB) and per-volume inference time on CPU and GPU (Tesla P100, 16 GB VRAM, Kaggle environment).

\section{RESULTS}
\subsection{Cross-Validation Results for BrainUNet with SSA Dataset}
 
\begin{figure}[h!]
    \centering
    \includegraphics[width=1\textwidth]{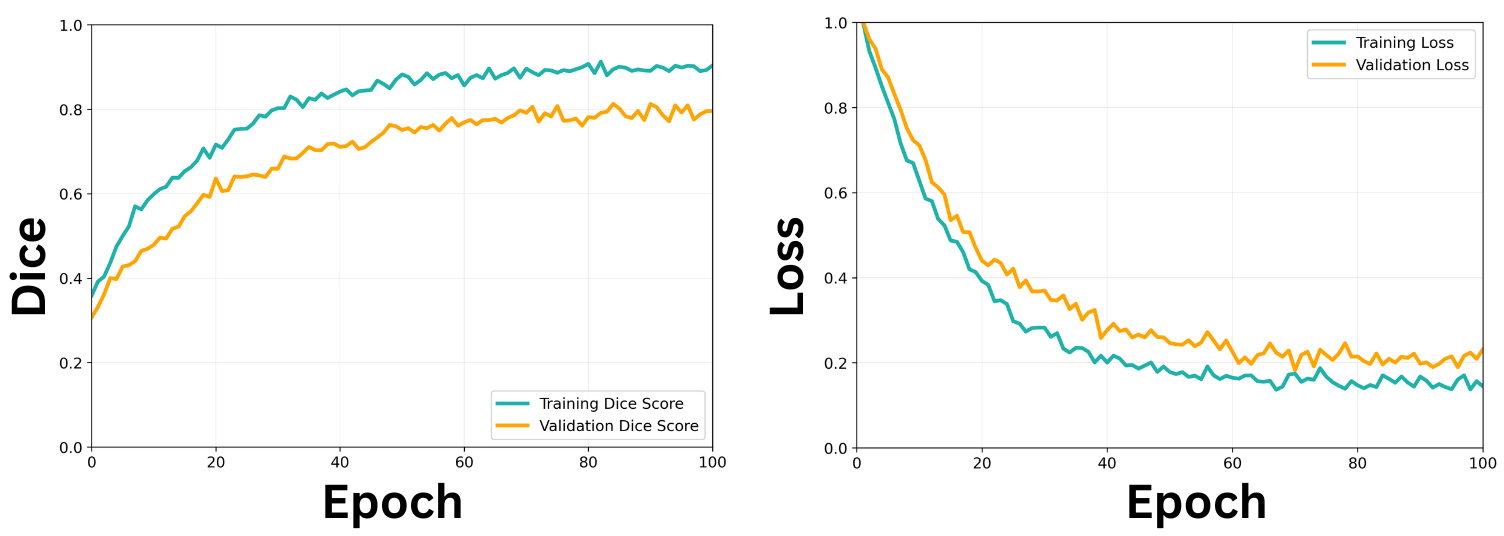}
    \caption{Average training and validation metrics across folds during cross-validation. (Left) Average Dice Score, and (Right) Average Loss. These figures reflect the performance of the BrainUNet model before fine-tuning.}
    \label{fig:crossval_metrics}
\end{figure}

The performance of the BrainUNet model before fine-tuning is summarized in Figures \ref{fig:crossval_metrics}, and shows the
average training and validation Dice Score, IoU, and Loss across the folds. The results indicated a
consistent improvement in training metrics over epochs, where the average training Dice score progressively
improved, reaching $0.7$, and the IoU increased to an average of $0.62$. The validation metrics (Dice score
stabilized at $0.55$ and IoU averaged $0.47$) showed a steady convergence, reflecting the model’s ability to
generalize across different data splits. Similarly, the loss curves exhibited steady convergence, with
training loss decreasing to $0.39$ and validation loss plateauing at $0.62$.

\vspace{-20pt}

\subsection{Synapse Evaluation: BrainUNet Before and After Fine-Tuning} 

\begin{figure}[h!]
    \centering    
    \includegraphics[width=0.6\linewidth]{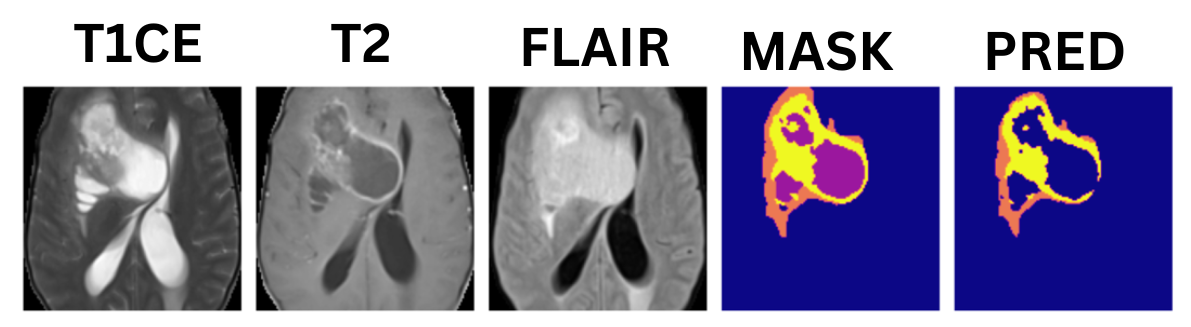}
    \caption{Segmentation results with BrainUNet before fine-tuning. T1CE, T2W, and FLAIR modalities are shown with ground-truth (Mask) and predicted segmentation (Pred).}
    \label{fig 4.jpg}
\end{figure}

We evaluated BrainUNet on the BraTS-Africa dataset before and after fine-tuning with BraTS-GLI 2021. As summarized in Table~\ref{tab:brainunet_results}, fine-tuning markedly improved Dice scores across all tumor subregions (ET, NETC, SNFH) and reduced Hausdorff distances, indicating more precise boundary delineation. Figure~\ref{fig 4.jpg} highlights a representative case, showing improved segmentation after fine-tuning.

\begin{table}[!h]
\centering
\begin{tabular}{|l|ccc|ccc|}
\hline
\textbf{Model} & \multicolumn{3}{c|}{\textbf{Dice Score}} & \multicolumn{3}{c|}{\textbf{Hausdorff Distance (95\%)}} \\ \hline
 & \textbf{ET} & \textbf{NETC} & \textbf{SNFH} & \textbf{ET} & \textbf{NETC} & \textbf{SNFH} \\ \hline
BrainUNet$_{SSA}$ & 0.52 & 0.47 & 0.62 & 74.8 & 77.48 & 23.78 \\
BrainUNet$_{SSA}$ fine-tuned & 0.76 & 0.80 & 0.85 & 30.00 & 30.55 & 18.72 \\ \hline
\end{tabular}
\vspace{6pt}
\caption{Performance of BrainUNet before and after fine-tuning, reported as Dice Score and Hausdorff Distance (95\%) for Enhancing Tumor (ET), Non-Enhancing Tumor Core (NETC), and Surrounding Non-Enhancing FLAIR Hyperintensity (SNFH).}
\label{tab:brainunet_results}
\end{table} 

To benchmark BrainUNet, we compared it against state-of-the-art baselines (nnUNet and MedNeXt). Results in Table~\ref{tab:brats_africa_comparison} show that although BrainUNet reports slightly lower Dice scores for ET and TC, it achieves competitive WT performance. Importantly, its compact size ($\sim$91 MB) and lower parameter count make it well suited for low-resource settings.

\begin{table}[h!]
\centering
\setlength{\tabcolsep}{4pt}
\renewcommand{\arraystretch}{1.1}
\begin{tabular}{|l|ccc|c|ccc|c|c|}
\hline
\textbf{Model} & \multicolumn{3}{c|}{\textbf{Lesion-wise Dice}} & \textbf{Avg} & \multicolumn{3}{c|}{\textbf{Legacy Dice}} & \textbf{Avg} & \textbf{Params} \\
               & \textbf{ET} & \textbf{TC} & \textbf{WT} &       & \textbf{ET} & \textbf{TC} & \textbf{WT} &       & \\
\hline
nnUNet          & 0.797 & 0.786 & 0.846 & \textbf{0.810} & 0.850 & 0.853 & 0.913 & \textbf{0.872} & 30M+ \\
MedNeXt Ens. & 0.793 & 0.786 & 0.893 & \textbf{0.824} & 0.878 & 0.876 & 0.924 & \textbf{0.893} & 50M+ \\
BrainUNet       & 0.684 & 0.714 & 0.831 & \textbf{0.743} & 0.759 & 0.791 & 0.869 & \textbf{0.806} & 22M+ \\
\hline
\end{tabular}
\vspace{7pt}
\caption{Lesion-wise and legacy Dice scores for BraTS-Africa segmentation. BrainUNet demonstrates competitive WT Dice while being significantly more lightweight (22M parameters, $\sim$91 MB), underscoring its suitability for deployment in resource-limited clinical environments.}
\label{tab:brats_africa_comparison}
\end{table}

\begin{figure}[htbp!]
    \centering
    \includegraphics[width=0.6\linewidth]{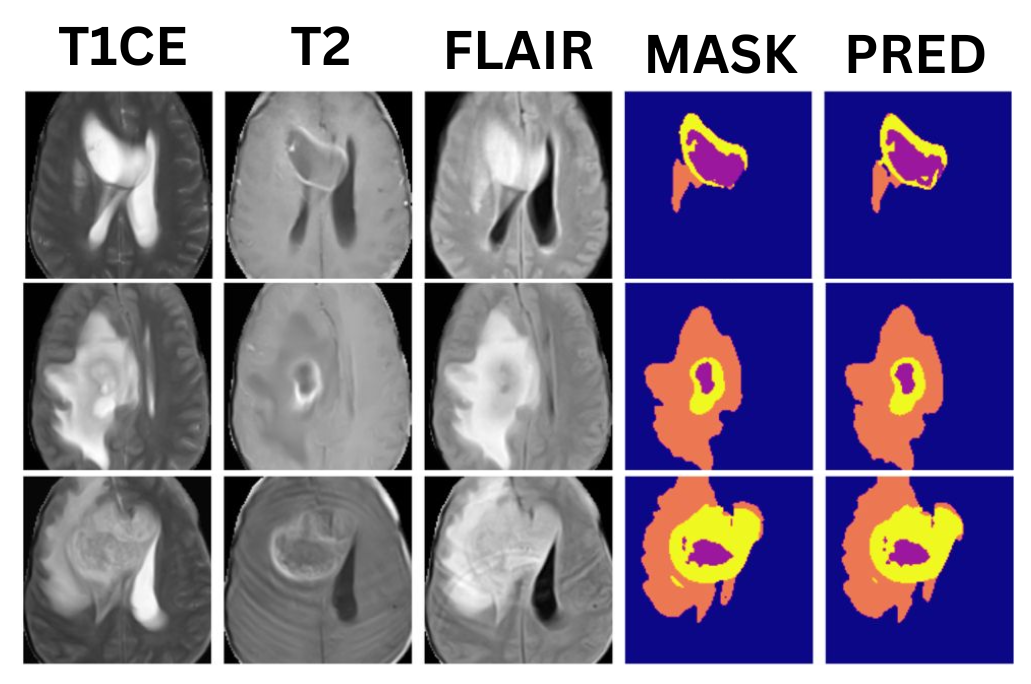}
    \caption{Final segmentation results from BrainUNet fine-tuned on BraTS-Africa validation data.}
    \label{fig:seg_tumor2}
\end{figure}

\subsection{Inference Analysis}
To assess the runtime performance of our model ($91.4$~MB), we measured per-volume inference time on both GPU and CPU. On a Tesla P100 (16~GB VRAM) in the Kaggle environment, the model processed each $3 \times 240 \times 240 \times 155$ voxel case in an average of 30.7~s. When run on CPU only (Intel\textsuperscript{\textregistered} Xeon\textsuperscript{\textregistered} @ 2.20~GHz), average time rose to 56.3~s per case. Despite the roughly $1.8\times$ slowdown, the sub-minute CPU runtime and compact weight file underline the model’s portability for low-resource settings. Table \ref{tab:inference_times} summarizes the detailed timings for the five validation cases.

\begin{table}[h!]
\centering
\begin{tabular}{|l|c|c|}
\hline
\textbf{Case ID} & \textbf{GPU (P100) Time (s)} & \textbf{CPU Time (s)} \\
\hline
BraTS-SSA-00126-000 & 30.69 & 55.82 \\
BraTS-SSA-00129-000 & 30.19 & 54.63 \\
BraTS-SSA-00132-000 & 30.28 & 55.43 \\
BraTS-SSA-00139-000 & 30.73 & 60.66 \\
BraTS-SSA-00143-000 & 31.41 & 54.80 \\
\hline
\textbf{Average} & \textbf{30.66} & \textbf{56.27} \\
\hline
\end{tabular}
\vspace{5pt}
\caption{Inference time per case on GPU (Tesla P10) and CPU (Intel Xeon @ 2.20GHz)}
\label{tab:inference_times}
\end{table}
\vspace{-35pt}
\section{Discussion and Conclusion}
\vspace{-5pt}
We introduced \textbf{BrainUNet}, a 3D U-Net variant enhanced with residual connections and attention mechanisms, tailored for glioma segmentation in sub-Saharan Africa (SSA). Using 5-fold cross-validation on the BraTS-Africa dataset, the model showed consistent improvements in Dice and IoU scores, though generalization to unseen samples remained challenging due to low resolution and imaging noise typical of SSA scans.

To address these limitations, we adopted a two-stage training strategy: pre-training on the large, high-quality BraTS 2021 dataset followed by fine-tuning on BraTS-Africa. This transfer learning approach improved segmentation performance, as validated on the Synapse platform using Dice scores and Hausdorff distances. While BraTS-Africa alone yielded reasonable results, fine-tuning substantially enhanced robustness against scanner variability and data artifacts.

In terms of deployment, BrainUNet demonstrated sub-minute inference times and compact memory size (91.4 MB), supporting feasibility in low-resource settings where GPU access is limited. These characteristics, combined with improved accuracy, highlight the model’s potential for clinical integration in LMICs.

Nonetheless, several limitations remain. The small size of the BraTS-Africa dataset may restrict generalizability across diverse SSA populations and scanner types. Persistent MRI quality issues, including motion artifacts and resolution variability, continue to affect prediction consistency. Runtime was tested only on moderate hardware, and broader evaluation on low-spec devices is needed. Finally, external validation was limited to a single dataset; multi-center studies would strengthen robustness claims. 

Future work should focus on enhancing interpretability, improving cross-dataset generalization, and exploring integration into real-time diagnostic workflows to maximize clinical impact.

\section*{Acknowledgments}

The authors gratefully acknowledge the faculty and instructors of the SPARK Academy 2024 Summer School for their invaluable insights on deep learning in medical imaging, particularly brain tumors. We also thank the Digital Research Alliance of Canada for computational infrastructure, the McGill University Doctoral Internship Program for student exchange support, and the BrainUNet team for their guidance during model development. This work was supported by the Lacuna Fund for Health and Equity (PI: Udunna Anazodo, grant \#0508-S-001), the Natural Sciences and Engineering Research Council of Canada (NSERC) Discovery Launch Supplement (PI: Udunna Anazodo, grant \#DGECR-2022-00136), and partly by the RSNA Research \& Education Foundation Derek Harwood-Nash International Education Scholar Grant (Investigators: Farouk Dako and Udunna Anazodo).   This work was partly supported by the Italian Ministry of University and Research (MUR) under project PE0000013 – Future of Artificial Intelligence Research (FAIR)

\begin{credits}

\subsubsection{\discintname}
The authors have no competing interests to declare that are relevant to the content of this article.
\end{credits}
%
%
%
\bibliographystyle{splncs04}
\bibliography{final_references}

\end{document}